\documentclass[sn-mathphys]{sn-jnl}

\jyear{2021}%

\theoremstyle{thmstyleone}%
\theoremstyle{thmstyletwo}%

\theoremstyle{thmstylethree}%
\newtheorem{definition}{Definition}%

\usepackage{subcaption}
\usepackage{amsmath}

\begin{document}

\title[Glocal Explanations of xG Models]{Glocal Explanations of Expected Goal Models in Soccer}


\author*[1]{\fnm{Mustafa} \sur{Cavus}}\email{mustafacavus@eskisehir.edu.tr}

\author[2]{\fnm{Adrian} \sur{Stańdo}}\email{adrian.stando.stud@pw.edu.pl}

\author[2]{\fnm{Przemysław} \sur{Biecek}}\email{przemyslaw.biecek@pw.edu.pl}

\affil[1]{\orgdiv{Department of Statistics}, \orgname{Eskisehir Technical University}, \orgaddress{\city{Eskisehir}, \postcode{26555}, \country{Turkiye}}}

\affil[2]{\orgdiv{Faculty of Mathematics and Information Science}, \orgname{Warsaw University of Technology}, \orgaddress{\street{ul. Koszykowa}, \city{Warsaw}, \postcode{00-662}, \country{Poland}}}


\abstract{ 
The expected goal models have gained popularity, but their interpretability is often limited, especially when trained using black-box methods. Explainable artificial intelligence tools have emerged to enhance model transparency and extract descriptive knowledge for a single observation or for all observations. However, explaining black-box models for a specific group of observations may be more useful in some domains. This paper introduces the glocal explanations (between local and global levels) of the expected goal models to enable performance analysis at the team and player levels by proposing the use of aggregated versions of the SHAP values and partial dependence profiles. This allows knowledge to be extracted from the expected goal model for a player or team rather than just a single shot. In addition, we conducted real-data applications to illustrate the usefulness of aggregated SHAP and aggregated profiles. The paper concludes with remarks on the potential of these explanations for performance analysis in soccer analytics.}




\maketitle

\section{Introduction}





In soccer, it is not uncommon for one team to dominate a match, creating many chances to score but failing to do so, while the opposing team manages to convert one of their few chances into a goal and win the match. Thus, the use of traditional end-of-match statistics is often argued against, because \textit{the number of shots}, \textit{ball possession percentage}, and \textit{shots inside the opponent's penalty area} do not always accurately reflect the outcome of the match. The rapid pace of technological advancements in data collection, storage, and analysis have had a revolutionary impact on soccer analytics over the last decade. Thanks to these advancements, soccer data is collected in two main forms: event data consists of ball-related events and where on the field they occurred such as \textit{shots}, \textit{passes}, \textit{tackles}, and \textit{dribbles} while tracking data consists of \textit{the position of players} and \textit{the ball throughout play on the pitch}. The technological revolution has made it possible to propose a large number of key performance indicators to measure different aspects of the game, such as \textit{pass evaluation}, \textit{quantification of controlled space}, \textit{shot evaluation}, and \textit{goal-scoring opportunities} using possession values. 

One of the most prominent metrics is the expected goal (xG), which has gained significant popularity within the soccer analytics society. Green \cite{green_2012} introduced the xG to estimate the probability that a shot will result in a goal, with the aim of providing a metric that accounts for the low-scoring nature of soccer, in contrast to other sports. In this context, xG serves as a valuable proxy for scoring in soccer. From a statistical point of view, it can be interpreted as the average of a considerable number of uncorrelated observations of the random variable corresponding to shots. In addition to being a reliable measure of scoring, xG has also been used as a predictor of future team performance \citep{cardoso_et_al_2019}. One interesting application story of the xG model is particularly at the club level in Denmark. The FC Midtjylland won their first Danish league title using the xG models to recruit players\footnote{\url{https://thecorrespondent.com/2607/How-data-not-humans-run-this-Danish-soccer-club/517995289284-77644562}}. They used the xG models to predict the future performance of the players and then made their recruitment decisions. There is certainly a number of different factors that may have contributed to this success, but it is a good example of the usefulness of the xG model in practice. Certainly, an xG model is not only used to recruit promising players. There are two main ways to use such a model: performance-based and ranking-based. In performance-based usage, if the calculated actual goals metric is lower than the xG, it means under-performance, and if on the other hand, it is higher, it indicates over-performance of a team or a player. The difference between the created xG and the allowed xG is the ranking-based usage of the xG metric which is used to measure and rank the performance of teams \cite{brechot_and_flepp_2020}. Due to the usefulness of the xG, there is much research in the literature on training an accurate xG model using both glass-box and black-box methods \cite{eggels_et_al_2016, pardo_2020, tippana_2020, anzer_and_bauer_2021, haaren_2021, umami_et_al_2021, fernandez_et_al_2021, cavus_and_biecek_2022, mead_et_al_2023}. However, the interpretability of models trained based on black-box methods is limited or impossible. Thus, it is not possible to debug, be accountable, and gather descriptive knowledge from the model.  

Explainable artificial intelligence (XAI) is an area that has gained a better understanding in recent years, aiming to make black-box machine learning (ML) models transparent, whose inner workings are difficult for end-users to understand. From finance \cite{bücker_et_al_2022} to medical sciences \cite{hryniewska_et_al_2021}, making such models more understandable in many different fields has helped to train ML models more responsibly rather than focusing on higher prediction accuracy. It also provides an opportunity to extract descriptive knowledge from the model. The XAI tools are generally classified into two levels: local and global explanations. While the local explanation tools are used to understand the model behavior at an observation level, the global explanation tools aim to understand it at the dataset or model level. The most popular of these tools are SHapley Additive exPlanations (SHAP) values, which can be used both at the local and global levels \cite{lundberg_and_lee_2017}. SHAP values are based on Shapley values in the cooperative game theory \cite{shapley_1953} and are used to measure the contribution of each feature to the model prediction for an observation. It can also be used for the dataset to measure the importance of the features in the model.  

There are some attempts to explain the black-box xG models in the soccer domain. Pardo \cite{pardo_2020}, Van Haaren \cite{haaren_2021}, and Mead et al. \cite{mead_et_al_2023} used the SHAP values to measure the importance of the features in the model. Bransen and Davis \cite{bransen_and_davis_2021} figured out the relationship between the features and the response variable by using the Partial Dependence Profiles. A common limitation in these studies is that the XAI tools were used only at the model level. However, it may be more useful to explain the model for a group of observations than simply explaining the model at the model or observation level. In this way, Cavus and Biecek \cite{cavus_and_biecek_2022} indicated that it is possible to extract knowledge from the model at the team or player level. They aggregated the ceteris-paribus profiles, which are used to examine the model behaviour at the local level for a variable, and measured the soccer player and team performances using expected goal models. The reason they use aggregated rather than individual profiles is that measuring player performance not on a single shot, but on all shots during the period of interest (e.g., a game, part of a season) provides more useful information in the domain of soccer analytic. However, the method proposed in this study can only be used to examine the relationship between a feature and the response variable, and it does not provide information on the contribution of each feature to the scoring probability. Thus, we propose to use the aggregation of SHAP values to decompose the xG models to make it possible to analyze the scoring potential of the team and player.

The main contributions of this paper can be summarised as follows: (1) we demonstrate how the local-level explanation XAI tools, such as SHAP and ceteris-paribus profiles, can be used by aggregating as glocal or semi-global level XAI tools, and (2) we provide examples of how the method can be used for analyzing the scoring potential of team and player in terms of expected goal models. The rest of the paper is structured as follows: first, we discuss the related works in the literature in Section~\ref{sec:related_works}, the methodology: the aggregated SHAP and aggregated profiles used in the paper are given in Section~\ref{sec:methodology}. Then, in order to show the usefulness of aggregated SHAP and aggregated profiles, the real data applications are conducted in Section~\ref{sec:app}, and concluding remarks are given in Section~\ref{sec:conclusions}.

\section{Related works} \label{sec:related_works}

This section discusses related work on xG models, explanations of xG models, and explanations of black-box ML models for a group of observations. 

\subsection{The xG model in soccer analytics}

Since xG has revolutionised soccer analytics in the last decade, it has been the subject of numerous scientific works. These studies can be divided into three parts: (1) proposing accurate xG models, (2) using xG models for performance evaluation, and (3) other works related to xG models.

Many xG models are trained using different strategies on the event data and both the event and tracking data to achieve better predictions. Eggels et. al. \cite{eggels_et_al_2016} and Fernandez et al. \cite{fernandez_et_al_2021} proposed a spatiotemporal features-based xG model. Herbinet \cite{herbinet_2018} conducted another study on the hybrid model which combined the xG model and ELO rating system to consider the current level of a team. Pardo \cite{pardo_2020} created an xG model using the qualitative player information. Wheatcroft and Sienkiewicz \cite{wheatcroft_and_sienkiewicz_2021} proposed a simple parametric model to predict both match outcome and the total number of goals which can outperform a model assuming an equal probability of shot success among teams. Umami et al. \cite{umami_et_al_2021} considered the joint effect of the features \texttt{distance to goal} and \texttt{angle to goal} in the model. Anzer and Bauer \cite{anzer_and_bauer_2021} used the hand-crafted features for proposing the xG model. On the other hand, Hewitt and Karakus \cite{hewitt_and_karakus_2023} extended the discussion about the features used in the xG models and they proposed a positional-adjusted xG model to get more accurate predictions. Mead et al. \cite{mead_et_al_2023} improved the performance of the xG model by using some unused features such as \textit{player ability} and \textit{psychological effects}. These studies followed basic strategies such as the use of different features, the use of models with different levels of complexity, and the size and partitioning of the data. A comprehensive methodological discussion of these strategies is made and important takeaway messages are given about training xG models in \cite{robberechts_et_al_2020}.

In addition to the studies focused on obtaining more accurate xG models, the second group of studies focuses on utilizing xG models for performance evaluation. Lucey et al. \cite{lucey_et_al_2014} proposed the \textit{quality-quantity approach} to measure the performance based on the xG model trained on spatiotemporal features. They compare xG values with actual goals, and if the actual goals are lower than the xG, it indicates under-performance, and if, on the other hand, it is higher, it indicates over-performance of a team or a player. Fairchild et al. \cite{fairchild_et_al_2018} approached player and team evaluation from the perspective of offensive and defensive efficiency by comparing xG with the actual goals. They developed an xG model for Major League Soccer in the USA and Canada. Brechot and Flepp \cite{brechot_and_flepp_2020} proposed the use of xG models for performance evaluation, emphasizing the potential influence of randomness on match outcomes in the short term. They introduced a chart that plotted teams' rankings in the league table against their rankings based on xG. Moreover, they proposed some useful metrics calculated based on xG such as offensive and defensive ratios. Kharrat et al. \cite{kharrat_et_al_2020} adapted the xG model with other most commonly used systems in soccer analytics such as \textit{plus-minus rating} to measure a player's contribution to the goal difference during the time player is on the pitch. Sarkar and Kamath \cite{sarkar_and_kamath_2021} used the difference between the actual and xG of teams to measure the variability of luck among the top and bottom six ranks and the determination of the rank positions. Toda et al. \cite{toda_et_al_2022} propose a method to evaluate team defense from a comprehensive perspective related to team performance by predicting ball recovery and being attacked using player actions and positional data of all players and the ball. The result of this combined system is also used to examine the potential transfer effect of players and to decide which player to recruit. Several variants of the xG are proposed and used in the performance evaluation such as xG against, non-penalty xG, non-penalty xG against, xG Chain, and xG Buildup\footnote{\href{https://statsbomb.com/articles/soccer/introducing-xgchain-and-xgbuildup/}{https://statsbomb.com/articles/soccer/introducing-xgchain-and-xgbuildup/}} with the increasing popularity of xG. Ruan et al. \cite{ruan_et_al_2022} aimed to identify and measure the effectiveness of different defensive playing styles for professional soccer teams considering the xGA in the Chinese soccer League. 

In addition to papers proposing accurate xG models and using xG models in performance evaluation, there are studies focusing on the recruitment process and a comparison of dynamics in the last group of papers. Spearman \cite{spearman_2018} proposed a probabilistic physics-based model that utilizes spatiotemporal player tracking data to quantify off-ball scoring opportunities. This model can be used in a number of different ways such as to obtain and analyze important positions during a match, to assist opposition analysis by highlighting the regions of the pitch where specific players or teams are more likely to create off-ball scoring opportunities, and to automate recruitment by finding the players across a league who are most efficient at creating off-ball scoring opportunities. Fernando et al. \cite{fernand_et_al_2015} utilized the xG model to compare the \textit{goal-scoring styles} of teams. Bransen and Davis \cite{bransen_and_davis_2021} conducted a comparison of the dynamics of men's and women's soccer in terms of \textit{goal-scoring rates} over the season, \textit{conversion rates}, and \textit{shot locations}. Raudonius and Seidl \cite{raudonius_and_seidl_2023} utilized the inherently interpretable xG model based on logistic regression to analyze the shooting tendencies, and efficiency, and explore how these change as players get older in German soccer leagues in terms of the model coefficients.

\subsection{Explanations of the xG models}

XAI tools are utilized to explain the black-box nature of machine learning models. The explanations of the xG model can be leveraged to enhance model performance through feature selection and extracting information for additional purposes such as performance evaluation and recruitment tasks. Papers related to the xG model explanations can be categorized into three groups: (1) interpreting the importance of features, (2) describing the relationship between the features and target variable in the xG model, and (3) using the explanations of the xG model for performance evaluation. Rathke \cite{rathke2017} found that the most important features in xG models are the \texttt{distance to goal} and \texttt{angle to goal}. Similarly, Pardo \cite{pardo_2020} and Van Haaren \cite{haaren_2021} investigated the importance of features using SHAP values and confirmed previous findings. Unlike these studies, Mead et al. \cite{mead_et_al_2023} explored the importance of unused features in the xG model in terms of SHAP values.

Bransen and Davis \cite{bransen_and_davis_2021} utilized Partial Dependence Profiles to examine the relationship between certain features and the response variable in the xG model for women's soccer. They also investigated whether an xG model developed for one gender can be applied to data from another gender and found that the same model is applicable. However, they observed some differences in the importance of features and how the models value certain types of shots. Cavus and Biecek \cite{cavus_and_biecek_2022} investigated the relationship between features such as \texttt{distance to goal}, \texttt{angle to goal}, and the target feature using the same XAI tool and profiled these relations. They also utilized these relations to compare the scoring potential of players. Thus, it may be possible to predict a player's potential goalscoring performance based on the features considered.

\subsection{Explanations of ML models for a group of observations} \label{subsec:exp}

In this section, we discuss the importance of XAI tools for explaining the behaviour of black-box ML models at a group level, rather than just for individual observations or all observations. While XAI tools are commonly categorized as \textit{local} and \textit{global} level explainers \cite{ema, molnar_2020, bhattacharya_2022}, recent studies have shown the need for explaining model behaviour for a specific group of observations in certain domains. For instance, in the context of xG models used in soccer, explaining the model's behaviour for a single shot (i.e., single observation) or all shots (i.e., all observations) may not provide valuable information for players or teams. Instead, focusing on explaining the model's behaviour for shots taken by a particular player or team can yield more insightful performance analysis \cite{cavus_and_biecek_2022}. 

Numerous studies across different domains demonstrate the usefulness of XAI tools for explaining model behaviour at a group level. One commonly used tool in this context is SHAP values, which are known for their additivity structure that allows for aggregation across multiple observations. For example, Berezo et al. \cite{berezo_et_al_2022} aggregated mean SHAP values for body parts to investigate how the location of certain wounds affects the model predictions. Bogatinovski et al. \cite{bogatinovski_et_al_2022} aggregated the absolute mean SHAP values to calculate token importance scores in natural language processing to alert developers during model construction. Kerr et al. \cite{kerr_et_al_2022} proposed aggregating absolute SHAP values at the city level to compare the importance of features for predicting air quality in different European cities. Pappalardo et al. \cite{pappalardo_et_al_2021} used the mean absolute SHAP values to compare the dynamics between men and women soccer teams on an AdaBoost model. On the other hand, Bowen and Ungar \cite{bowen_and_ungar_2020} introduced generalised SHAP to extract additional knowledge from the model by using SHAP values, such as classification explanation, group differences in model predictions, and model failure. Kruse et al. \cite{kruse_et_al_2021a, kruse_et_al_2021b} proposed using daily aggregated SHAP values to explore daily or seasonal trends in seasonal prediction problems such as weather or traffic. Laberge et al. \cite{laberge_et_al_2021} discussed mean aggregation of SHAP values and its challenges, while Mase et al. \cite{mase_et_al_2022} proposed a Bayesian bootstrap approach to measuring model fairness for underrepresented groups in society using individual and aggregated SHAP values. Matthews and Hartman \cite{matthews_and_hartman_2022} introduced multiplicative SHAP values for two-part models, commonly used in actuary, and provided a framework for their calculation. These studies exemplify the versatility and broad applicability of SHAP values.

In addition to SHAP values, aggregated Ceteris-Paribus profiles have been used to explain model behaviour for a group of observations. Cavus and Biecek \cite{cavus_and_biecek_2022}  employed aggregated profiles, referred to as semi-global explanations, to extract valuable descriptive information from black-box models. Although these XAI tools are used to explain model behaviour for groups of observations, to the best of our knowledge, a formal definition of this level of explanation is currently lacking. Therefore, we propose a new level of XAI tools to fill this gap in Section~\ref{sec:methodology}.

\section{Metholodology: Glocal Explanations} \label{sec:methodology}

The XAI tools are generally classified under two groups: (1) \textit{local explanation tools} on a prediction level, and (2) \textit{global explanation tools} on the model level \cite{ema, molnar_2020, bodria_et_al_2021}. Let's assume $X \in \mathbb{R}^d$ represent $d$-dimensional feature space, and $Y \in \{0, 1\}$ represent the binary target space. A classification model aims to learn a prediction function $f: X \to Y$. We use $(X_1, X_2, ..., X_d)$ and $Y$ to denote the random variables associated with the feature and target spaces, respectively, which are part of the joint data distribution $P(X, Y)$. A dataset $D = \{(x_i, y_i)\}^{n}_{i=1}$ consists of $n$ samples drawn independently and identically distributed from $P(X, Y)$. The $i$-th observation is denoted as $\textbf{x}_i = (x_{i}^1, x_{i}^2, ..., x_{i}^d)$, and the realizations of the $j$-th feature $X^j$ are denoted as $\textbf{x}^j = (x_{1}^j, x_{2}^j, ..., x_{n}^j)$. When the model $f$ is black-box, it can be explained by using any explainer function $e$. Thus, any local ($e_L$) and global explanation ($e_G$) of a model can be defined as in Eq.~(\ref{eq:1}) and (\ref{eq:2}), respectively

\begin{equation} \label{eq:1}
    e_L[f(X), \textbf{x}_i] = e_L(f, \textbf{x}_i),
\end{equation}

\begin{equation} \label{eq:2}
    e_G[f(X), D] = e_G(f, \{(x_i, y_i)\}^{n}_{i=1}).
\end{equation}\vspace{1mm}

In recent years, some papers use other categories to classify these tools. The term \textit{glocal} has been used for first time by Setzu et al. \cite{setzu_et_al_2021} to call their proposed method GlocalX which generates global explanations for a black-box model by hierarchically aggregating similar local explanations \cite{li_et_al_2022, alkhatib_et_al_2022, mahya_and_johannes_2023}. Achtibat et al. \cite{achtibat_et_al_2022} identified glocal XAI, which aims to combine local and global XAI perspectives in order to enhance explainability by minimising the observer's interpretation workload. Dreyer et al. \cite{dreyer_et_al_2023} indicated that glocal XAI methods strive to bridge the gap between global-scale concept visualization and the attribution of their significance in individual model inferences per sample. The usage of the term \textit{glocal} in these studies refers to the approach introduced by Ljungberg et al. \cite{lundberg_et_al_2020}, which involves explaining the global behaviour of the model through an aggregation of local-to-global analysis strategies. The aggregation strategies can be identified in three ways: (1) \textit{data aggregation}: Transform the dataset into a subset that consists of an interested group of observations, train a model on the subset, and explain it, (2) \textit{prediction aggregation}: Train a model on the dataset and aggregate the prediction for the subset, and (3) \textit{explanation aggregation}: Train a model on the dataset, explain the observations of interest and aggregate these explanations. We follow the third way, and suggest using the term \textit{glocal} to describe the behaviour of the model for a given set of observations, as in Eq.~(\ref{eq:3})

\begin{equation} \label{eq:3}
    e_{GL}[f(X), M] = e_{GL}(f, \{(x_i, y_i)\}^{m}_{i=1}),
\end{equation}\vspace{1mm}

\noindent where a group of observations $M = \{(x_i, y_i)\}^{m}_{i=1}$ for $m < n$. This way of model explanation can be referred to as glocal explanation which can be also called dataset-wise level similar to Wagner et al. \cite{wagner_et_al_2023}. They introduce attribution maps that are aggregated over entire subgroups of patients and propose the computation of aggregated beat-aligned attributions across subgroups with shared pathologies as a means to infer global model insights. 

\begin{figure}[H]
    \centering
    \includegraphics[trim = 3cm 17cm 3cm 17cm, clip = true, scale = 0.33]{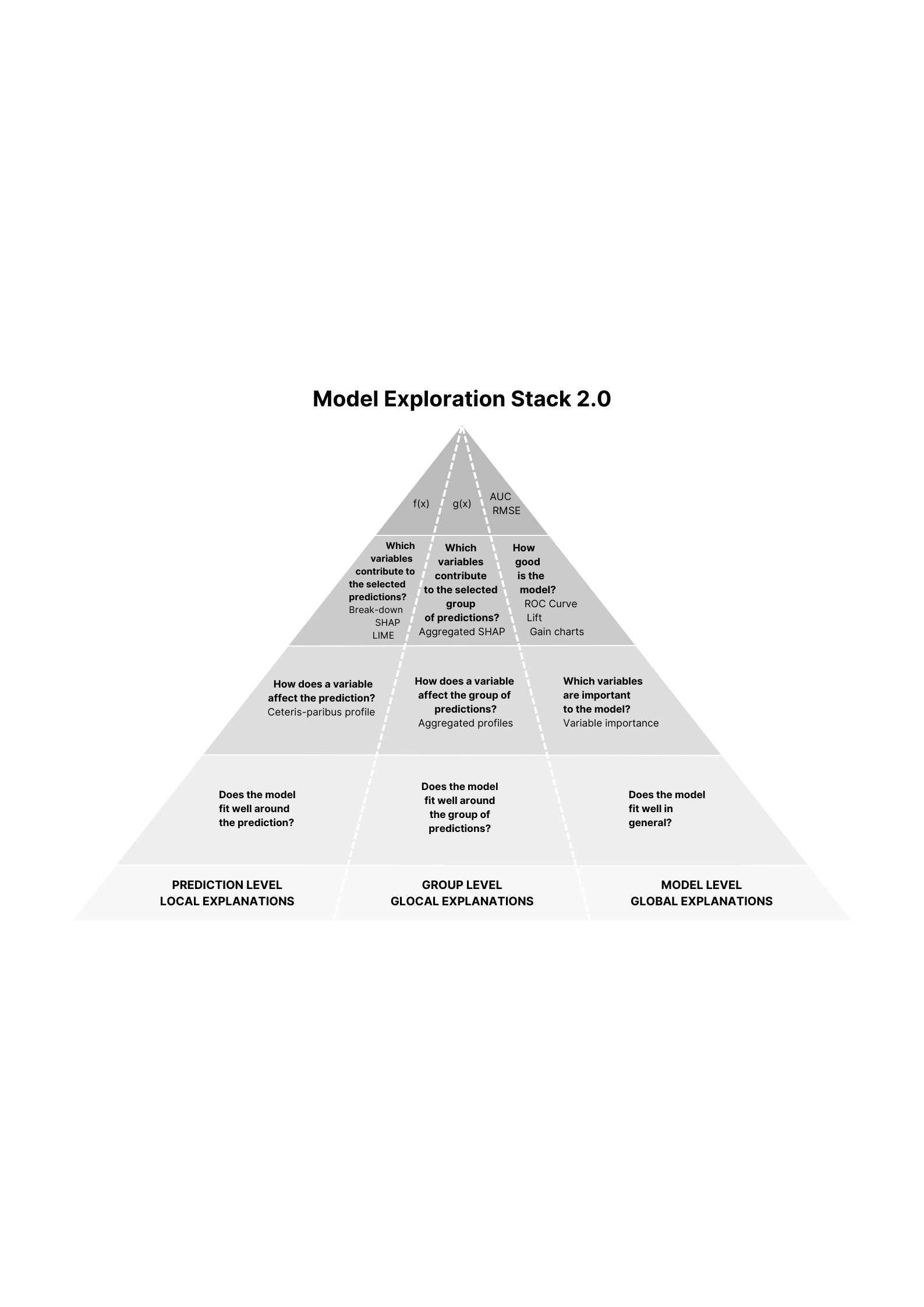}
    \caption{The extended version of the model exploration stack in \cite{ema} by adding the \texttt{Glocal explanation} component.}
    \label{fig:stack}
\end{figure}

As discussed in Section~\ref{subsec:exp}, an explanation of the black-box model may be more useful for a group of observations in several domains. However, to the best of our knowledge, there is no structurally defined level of XAI tools for this purpose. For this reason, we introduce a new section of XAI tools as \emph{glocal explanation tools} to explain the black-box model for a group of observations. Moreover, we introduce the \emph{aggregated SHAP} with the mathematical background and the \emph{aggregated profiles} from \cite{ema} as glocal explanation tools in Sections~\ref{subsec:aSHAP} and \ref{subsec:aProfiles}. These tools aim at answer the following questions about the model at the glocal level can be also seen in Fig.~\ref{fig:stack}.

\begin{enumerate}
    \item Which variables contribute to the selected group of predictions?
    \item How does a variable affect the group of predictions?
\end{enumerate}

\subsection{Aggregated SHAP} \label{subsec:aSHAP}

The Shapley Additive Explanations (SHAP) have been introduced as a method based on the Shapley values \cite{shapley_1953} in game theory for explaining model behaviour \cite{lundberg_and_lee_2017}. These values decompose the model prediction into separate components that can be explicitly attributed to individual variables. This allows for a clear understanding of the contribution of each variable to the model prediction

\begin{equation}
    f(X) = \phi_0 + \sum_{j = 1}^k \phi_j,
\end{equation}

\noindent where $X$ is the vector of $k$ variables and $f(X)$ is the prediction from the model for this vector. The $\phi_j$ term corresponds to the additive component for variable $j$, providing a measure of its contribution to the prediction made by the model at point $X$. The expected value, $\phi_0$, also known as the intercept, represents the baseline prediction of the model.

The purpose of this approach is to evaluate how the presence or absence of variable $j$ impacts the model prediction for a given instance, relative to the average prediction. To accomplish this, the original model predictions are compared when variable $j$ is included and excluded from the prediction

\begin{equation}
    \phi_j(f) = \sum_{k \in \{1, 2, ..., p\} / \{j\}} \frac{\lvert k\rvert ! (p - \lvert k\rvert- 1)!}{p!} [f(x_k)],
\end{equation}

\noindent where $x_j$ is the value of variable $j$, $k$ is the subset of variables, and $p$ is the number of variables in the model. In practice, $f(x_K)$ is estimated by substituting values for the remaining variables ${x_1, x_2, \ldots, x_p}\setminus k$ from a randomly selected observation. Each SHAP value, denoted by $\phi_j$, represents the difference between $f(x_i)$ and $f(z)$ due to variable $j$. 

Because the properties of Shapley values, including \textit{symmetry}, \textit{additivity}, and \textit{local accuracy}, are also applicable to predictive models \cite{ema}. Furthermore, based on the principles of \textit{additivity} and \textit{local accuracy}, the SHAP values can be aggregated \cite{kruse_et_al_2021a, kruse_et_al_2021b}. It means that SHAP values can be used as local explanations and they can be aggregated into global explanations \cite{doumard_et_al_2022}. Here we aggregate the SHAP values, as they are calculated per shot, leveraging their \textit{additivity} property to assess the scoring potential of a team and player. To obtain the SHAP values for observation (e.g. shot) $i$, the following formula can be used

\begin{equation}
    f_i(X) = \phi_0 + \sum_{j = 1}^{p} \phi_{ji},
\end{equation}

\noindent where $p$ is the number of variables and $\phi_{ji}$ is the SHAP value of variable $j$ for observation $i$. The sum of the SHAP values $\phi_{ji}$ for variable $j\ =1,\ \ 2,\ ...,\ p$, and a group of observations (e.g. shots of a team/player) $i\ =\ 1,\ 2,..., n$, the aggregated SHAP (aSHAP) values $f_A (X)$ can be given as follows

\begin{equation}
    f_{A}(X) = \phi_0 + \sum_{i = 1}^{n}\sum_{j = 1}^{p} \phi_{ji},
\end{equation}

\noindent where $n$ is the number of aggregated observations. The aSHAP values are calculated for the desired group of observations. This method is implemented in \texttt{shapviz} package \cite{shapviz} of \texttt{DALEX} XAI ecosystem \cite{dalex}.

\subsection{Aggregated profiles} \label{subsec:aProfiles}

The aggregated profiles (AP) are introduced in \cite{ema}. The idea behind the AP is the aggregation of the ceteris-paribus (CP) profiles that show how the change of a model’s prediction would change with respect to the value of a feature. In other words, a CP profile is a function that describes the dependence of the conditional expected value of the response with respect to the feature $j$. The AP can be defined simply as the averaging of the CP profiles that are considered. The value of an AP for model $f(.)$ and feature $j$ is defined as follows

\begin{equation}
    g_{AP}^j (z) = E_{\textbf{X}}^{-j}[f(\textbf{X}^{-j\mid=z})],
\end{equation}

\noindent where $g_{AP}$ is the expected value of the model predictions when $X_j$ is fixed at $z$ over the marginal distribution of $\textbf{X}_{j \mid z}$. The distribution of $\textbf{X}_{j \mid z}$ can be estimated by using the mean of CP profiles for $X_j$ as an estimator of the AP

\begin{equation}
    \hat{g}^j_{AP} (z) = \frac{1}{k} \sum_{i = 1}^{k}f(\textbf{x}^{ij \mid z}),
\end{equation}

\noindent where $k$ is the number of profiles that are aggregated. The difference between the AP and PDP is the number of aggregated profiles. The PDP is the aggregation of all profiles which are calculated on the entire data set while the AP is the aggregation of a group of profiles. 

\section{Applications} \label{sec:app}

The xG is a reliable indicator for performance analysis because of the high correlation with player performance \cite{davis_et_al_2022}. In this section, we show how glocal explanations of the xG model can be used in performance analysis through several use cases.

We used the pre-trained xG model from \cite{cavus_and_biecek_2022} because it is the best-performing of all the alternatives. It has been trained on event data consisting of 315,430 shots from 12,655 matches played in the German Bundesliga, English Premier League (EPL), Spanish La Liga, France Ligue 1, and Italy Serie A during the seasons between 2014-15 and 2020-21 with the features \texttt{minute}, \texttt{homeAwayTeam}, \texttt{situation}, \texttt{shotType}, \texttt{lastAction}, \texttt{distanceToGoal}, and \texttt{angleToGoal} from Understat\footnote{https://understat.com}. The details of the variables are given in Table~\ref{tab:variables}.

\begin{table}[htb]
    \centering
    \caption{The details of the variables used in the xG model}
    \label{tab:variables}
    \begin{tabular}{p{2.5cm}p{2cm}p{6cm}}\hline
    
        Variable                    & Type          & Description \\\hline
        \texttt{angleToGoal}      & continuous    & angle of the throw to the goal line \\
        \texttt{distanceToGoal}   & continuous    & distance from where the shot was taken to the goal line\\
        \texttt{shotType}          & categorical   & type based on the limb used by the player to shoot (\texttt{Head}, \texttt{Left foot},
\texttt{Right foot}, \texttt{another part of the body})\\
        \texttt{situation}          & categorical   & situation at the time of the event (\texttt{Direct freekick}, \texttt{From corner},
\texttt{Open play}, \texttt{Penalty}, \texttt{Set play})\\
        \texttt{homeAwayTeam}      & categorical   & status of the shooting team (\texttt{home} or \texttt{away})\\
        \texttt{lastAction}        & categorical   & last action before the shot (\texttt{Pass}, \texttt{Cross}, \texttt{Rebound}, \texttt{Head Pass}, and
35 more levels)\\
        \texttt{minute}             & continuous    & minute of shot\\\hline
    \end{tabular}
    
\end{table}

The aggregated profiles are used to analyze the scoring potential of a player in Section~\ref{sec:score_potential} and the blind spots of a goalkeeper's performance in Section~\ref{sec:goalkeeper_analysis}. The aSHAP is used to analyze the performance changes of a team in Section~\ref{sec:performance_change}.

\subsection{Scoring potential analysis with the aggregated profiles}\label{sec:score_potential}

The contribution of players to the creation of chances is a matter of curiosity in soccer analytics \cite{bransen_et_al_2019, decroos_et_al_2021}. Therefore we focus on measuring the potential contribution of the young players. Here we focus on the most valuable under 18 age players from top-5 European leagues which is collected from Transfermarkt\footnote{\href{Tranfermarkt - The list of the most valuable under 18 age players from top-5 European leagues}{https://www.transfermarkt.com.tr/spieler-statistik/wertvollstespieler/marktwertetop/plus/0/galerie/0?ausrichtung=Sturm&spielerposition_id=alle&altersklasse=u18&jahrgang=0&land_id=0&kontinent_id=0&yt0=G\%C3\%B6ster}} in Table~\ref{table:u18}.

\begin{table}[htb]
    \centering
    \caption{The shot and goal statistics of the most valuable under 18 age players from top-5 European leagues in the season of 2022/23}
    \begin{tabular}{lllcc}\hline
    Player              & Team                      & League       & Shots & Goals \\\hline
    Youssoufa Moukoko   & Borussia Dortmund         & Bundesliga   & 35    & 7  \\
    Alejandro Garnacho  & Manchester United         & EPL          & 24    & 3  \\
    Mathys Tel          & Bayern Munich             & Bundesliga   & 20    & 5  \\
    Jamie Bynoe-Gittens & Borussia Dortmund         & Bundesliga   & 24    & 3  \\
    Evan Ferguson       & Brighton \& Hove Albion   & EPL          & 36    & 6  \\\hline
    \end{tabular}
    \label{table:u18}
\end{table}

\begin{figure}[htb]
    \centering
    \includegraphics[width=0.75\linewidth]{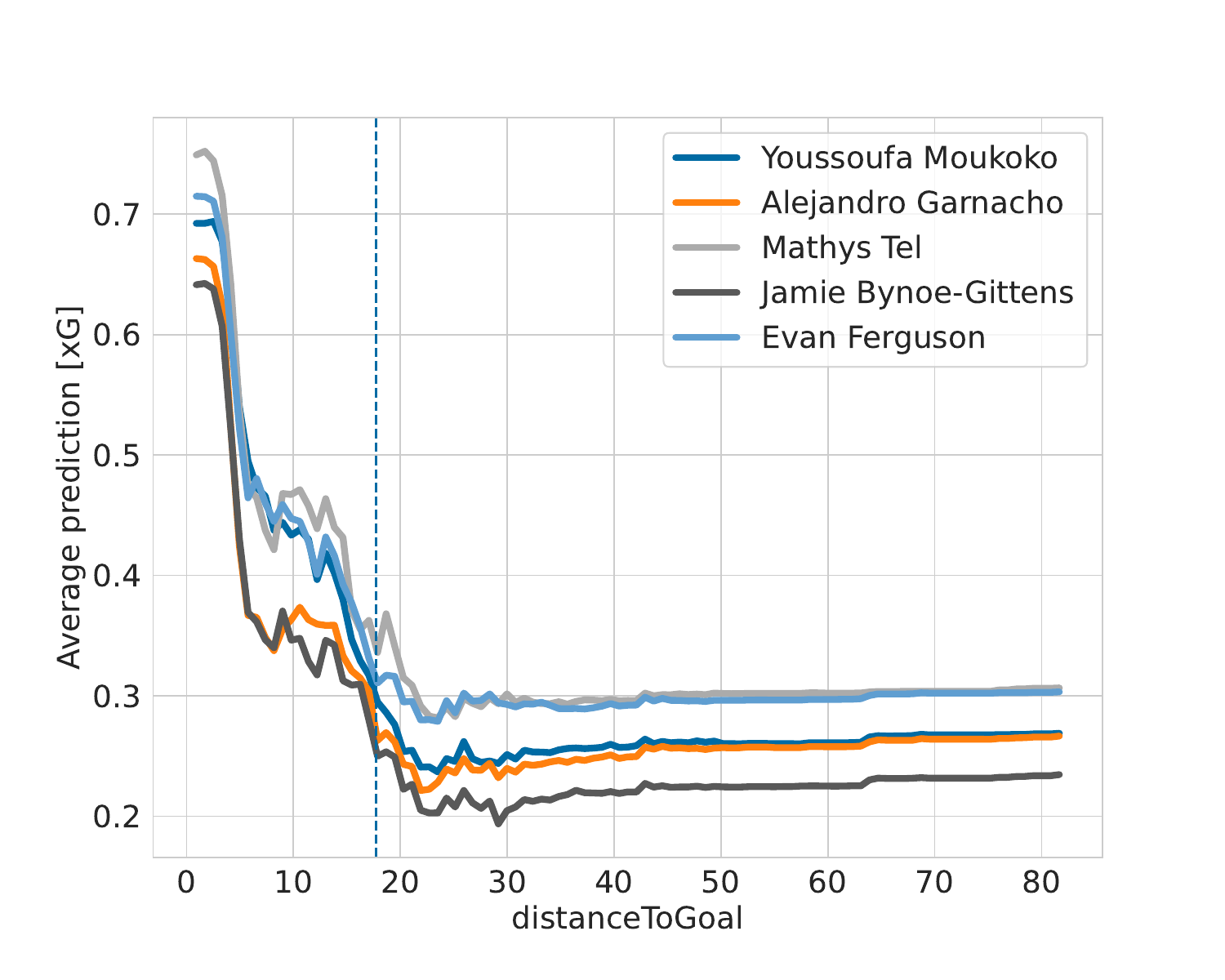}
    \caption{Aggregated profiles of the young players' scoring potential for the variable \texttt{distanceToGoal}. The dashed vertical line indicates the mean observation of the variable for all players.}
    \label{fig:yp-ap-distance}
\end{figure}

\begin{figure}[htb]
    \centering
    \includegraphics[width=0.75\linewidth]{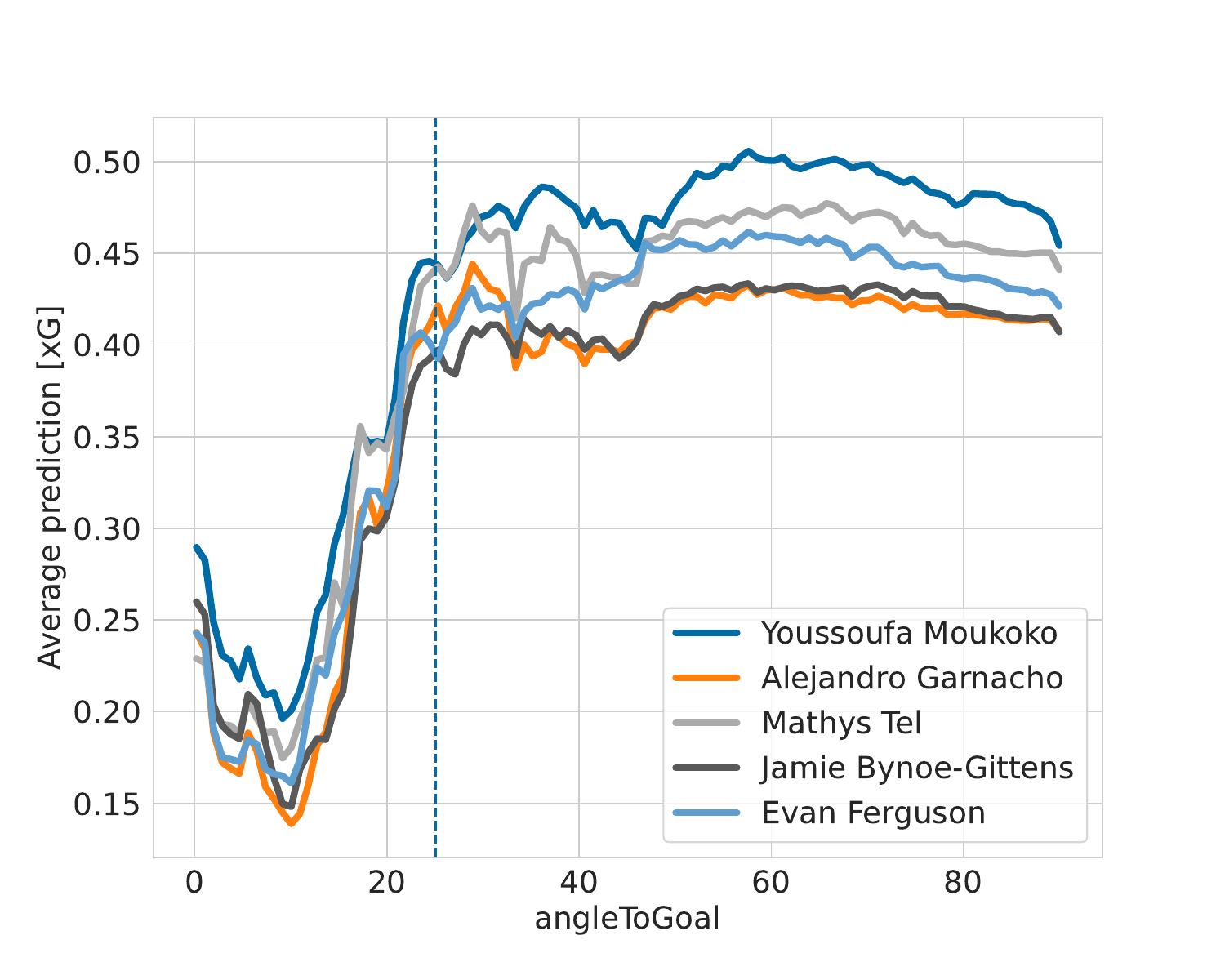}
    \caption{Aggregated profiles of the young players' scoring potential for the variable \texttt{angleToGoal}. The dashed vertical line indicates the mean observation of the variable for all players.}
    \label{fig:yp-ap-angle}
\end{figure}

The aggregated profiles of the players which are calculated regarding the shots of the players during the season of 2022/23 for the most important two variables \texttt{distanceToGoal} and \texttt{angleToGoal} in the xG model are given in Fig~\ref{fig:yp-ap-distance} and \ref{fig:yp-ap-angle}, respectively.

In Fig~\ref{fig:yp-ap-distance}, the scoring potential of the players can be evaluated under two groups for $0-15$ values of \texttt{distanceToGoal} according to the similarity of their profiles: the better group (higher average prediction is better) consists of \textbf{Youssoufa Moukoko}, \textbf{Evan Ferguson}, \textbf{Mathys Tel}, and the rest of the players in another group. Thus, the players in the better group can be evaluated as having higher scoring potential at a relatively closer distance. When the values of \texttt{distanceToGoal} are higher than $20$, the scoring potential of \textbf{Youssoufa Moukoko} approximates the worse group. Consequently, it is evaluated that the scoring potential of \textbf{Evan Ferguson} and \textbf{Mathys Tel} is better than the others in terms of aggregated profiles that are created to explain the xG model for players.

Unlike in Fig~\ref{fig:yp-ap-distance}, the scoring potential of the players could not be evaluated among the groups for the values of \texttt{angleToGoal} according to the similarity of their profiles. The scoring potential of the players is similar on the harder angles (lower is harder which means not in the front of the goal). However, the better players are \textbf{Youssoufa Moukoko}, \textbf{Mathys Tel}, and \textbf{Evan Ferguson}, respectively on the easier angles. It is seen that \textbf{Jamie Bynoe-Gittens} and \textbf{Alejandro Garnacho} have lower scoring potential. The findings from Fig~\ref{fig:yp-ap-distance} and \ref{fig:yp-ap-angle} show that the scoring potential of \textbf{Mathys Tel} is slightly better while \textbf{Jamie Bynoe-Gittens} and \textbf{Alejandro Garnacho} do not have the competitive potential with the others.

\subsection{Blind spot analysis of the goalkeepers' performance with the aggregated profiles}\label{sec:goalkeeper_analysis}

A goalkeeper has many responsibilities, such as distributing the ball and communicating with defenders \cite{jamil_et_al_2021}, but the main one is to save the shots on target. The goalkeeper's performance can be easily assessed by looking at the ratio of saves to shots \cite{gelade_2014}. On the other hand, the expected save (\textbf{xS}) metric, which measures the probability that a goalkeeper will save the shot, is proposed to evaluate a goalkeeper's performance, similar to the \textbf{xG} metric\footnote{\href{https://deepxg.com/2015/10/20/expected-saves/}{https://deepxg.com/2015/10/20/expected-saves/}}. The \textbf{xS} metric is the complement of the \textbf{xG} metric, i.e., $\textbf{xS} = 1 - \textbf{xG}$, but the only difference is that shots on target are taken into account instead of all shots  \cite{ruiz_et_al_2017}. 

In this section, instead of training an xS model, we use the xG metric to proxy ta goalkeeper's performance in terms of expected goal against on-target (\textbf{xGAOT}) which is the combination of the expected goal against (\textbf{xGA}) and expected goal on-target (\textbf{xGOT}). These metrics are the variants of the \textbf{xG} metric, which is given in Definition \ref{def}.

\begin{definition}[The \textbf{xG} variants] \label{def}
Let $\mathbf{X}$ be the explanatory variables used to predict the expected goal value $y_i$ for the shot $i\in \{1, 2, ..., n\}$ in the expected goal model $f$. The $\hat{y}_i = f(\mathbf{X}_i)$ is the predicted expected goal value for a shot $i$, and the $z_i\in \{0, 1\}$ is a binary variable that represents the shot $i$ on target or not. It takes the value of $1$ if the shot $i$ is on-target, and $0$ if not. The expected goal value is calculated as \textbf{xG} $= \sum_{i=1}^m \hat{y}_i$ for $m$ shot(s) where ($m < n$). The expected goal on-target is calculated as \textbf{xGOT} $= \sum_{i=1}^m y_i z_i$ which is for the prediction of shots on-target. Conversely, the expected goal against is calculated as \textbf{xGA} $=\sum_{l=1}^m \hat{y}_l$ for the conceded shots $l = 1, 2, ..., m$ by the opponent team. Like \textbf{xGOT}, the expected goal against on-target is calculated as \textbf{xGAOT} $=\sum_{l=1}^m y_l z_l$ for the conceded shots on-target.
\end{definition}

Each variant provides the ability to evaluate player or team performance from different perspectives. The \textbf{xGA} can be employed to assess the defensive performance of a team or players. A higher \textbf{xGA} value indicates that the shots faced by the team or player have a higher potential of resulting in a goal. This might suggest a weak defense or that the opposing team is creating more dangerous scoring opportunities. This metric can be valuable for analyzing team performance, reviewing defensive strategies, and identifying weaknesses. By analyzing \textbf{xGA} values, teams can develop strategies to strengthen their defense or make the opponent's shots less effective. Here, we use the \textbf{xGAOT} metric, which shows the expected goal value that a goalkeeper faced the shot on-target, to analyze the blind spots of the goalkeeper's performance. 

We have considered three goalkeepers under the age of 30 who have played in all league matches from three different leagues listed in Table~\ref{tab:goalkeeper}. The reason we focus on goalkeepers playing in all matches and those under the age of 30 are to eliminate the bias created by the difference in difficulty levels of the opponent teams, and to consider the age effect on the percentage of shots-on-target saves \cite{raudonius_and_seidl_2023}, respectively.

\begin{table}[htb]
    \centering
    \caption{The basic information and statistics of three goalkeepers playing in all matches and those under the age of 30}
    \begin{tabular}{lccll}\hline
    Goalkeeper      & Age   & Conceded Goals    & Team          & League \\\hline 
    Marvin Schwabe  & 27    & 54                & FC Köln       & Bundesliga \\
    Alex Remino     & 27    & 69                & RCD Espanyol  & La Liga\\
    David Raya      & 26    & 46                & Brentford FC  & EPL \\\hline
    \end{tabular}
    \label{tab:goalkeeper}
\end{table}

\begin{figure}[htb]
    \centering
    \includegraphics[width=0.75\linewidth]{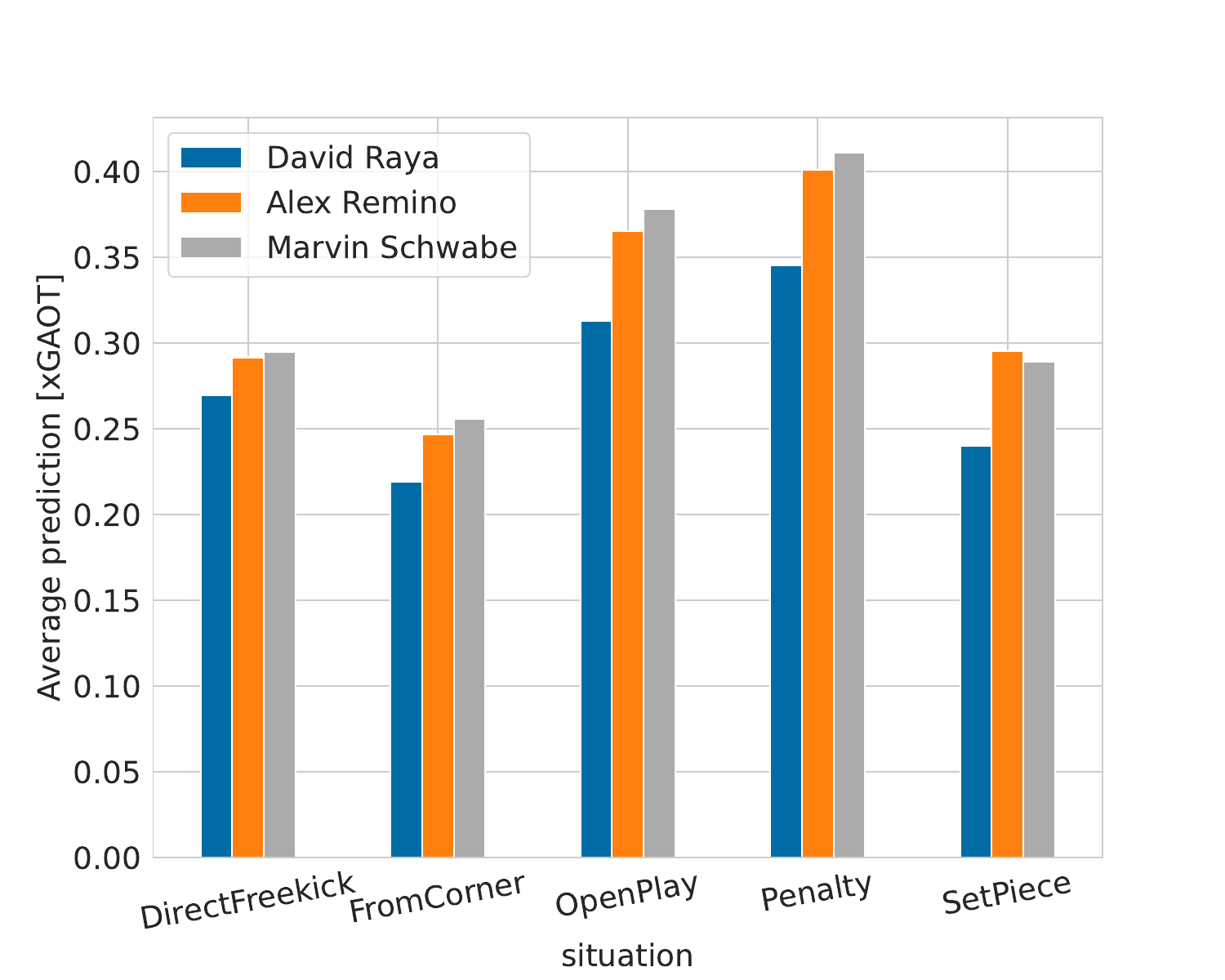}
    \caption{Aggregated goalkeeper blind spot profiles for the variable \texttt{situation}.}
    \label{fig:gk-ap-situation}
\end{figure}

\begin{figure}[htb]
    \centering
    \includegraphics[width=0.75\linewidth]{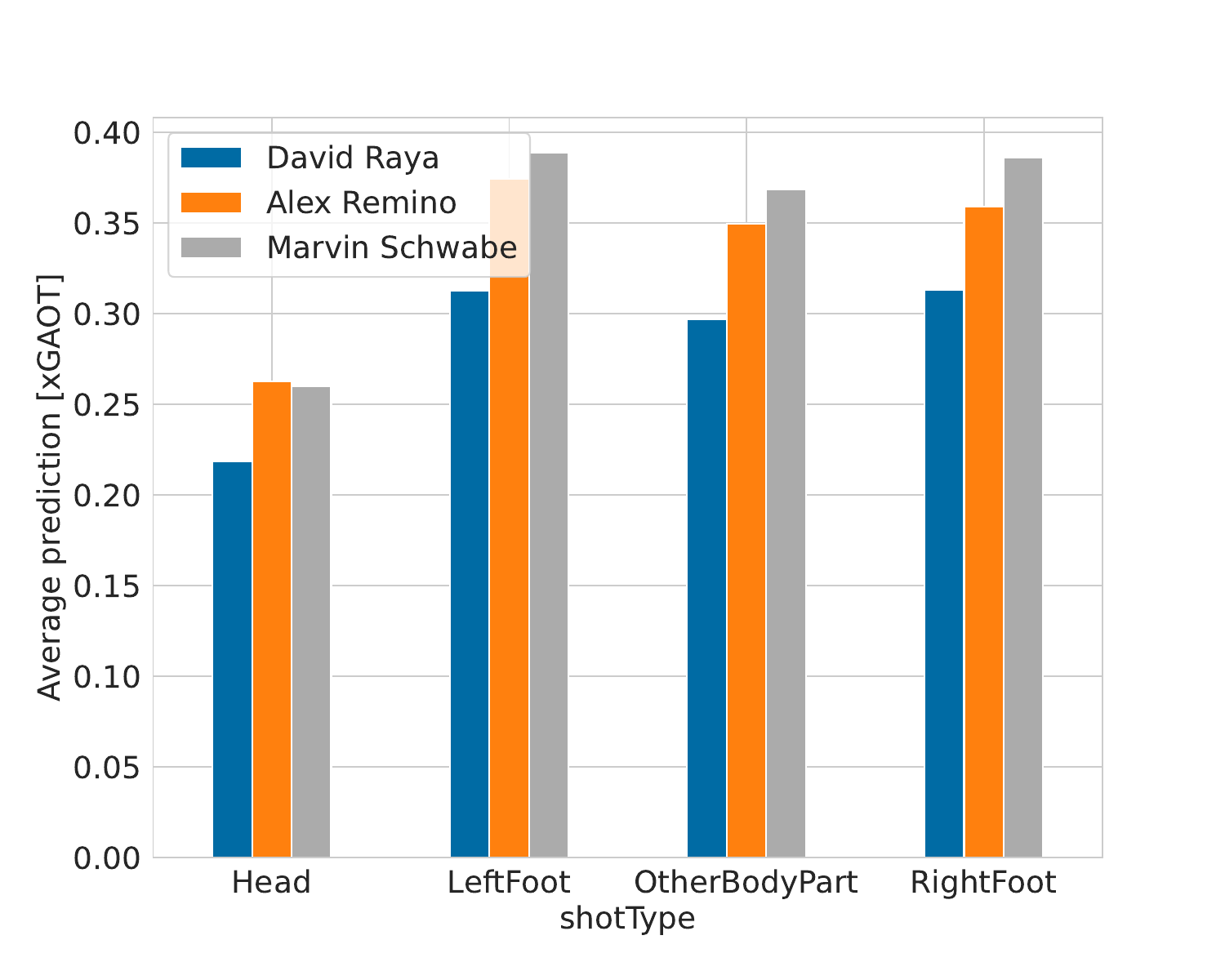}
    \caption{Aggregated goalkeeper blind spot profiles for the variable \texttt{shotType}.}
    \label{fig:gk-ap-shot}
\end{figure}

\begin{figure}[htb]
    \centering
    \includegraphics[width=0.75\linewidth]{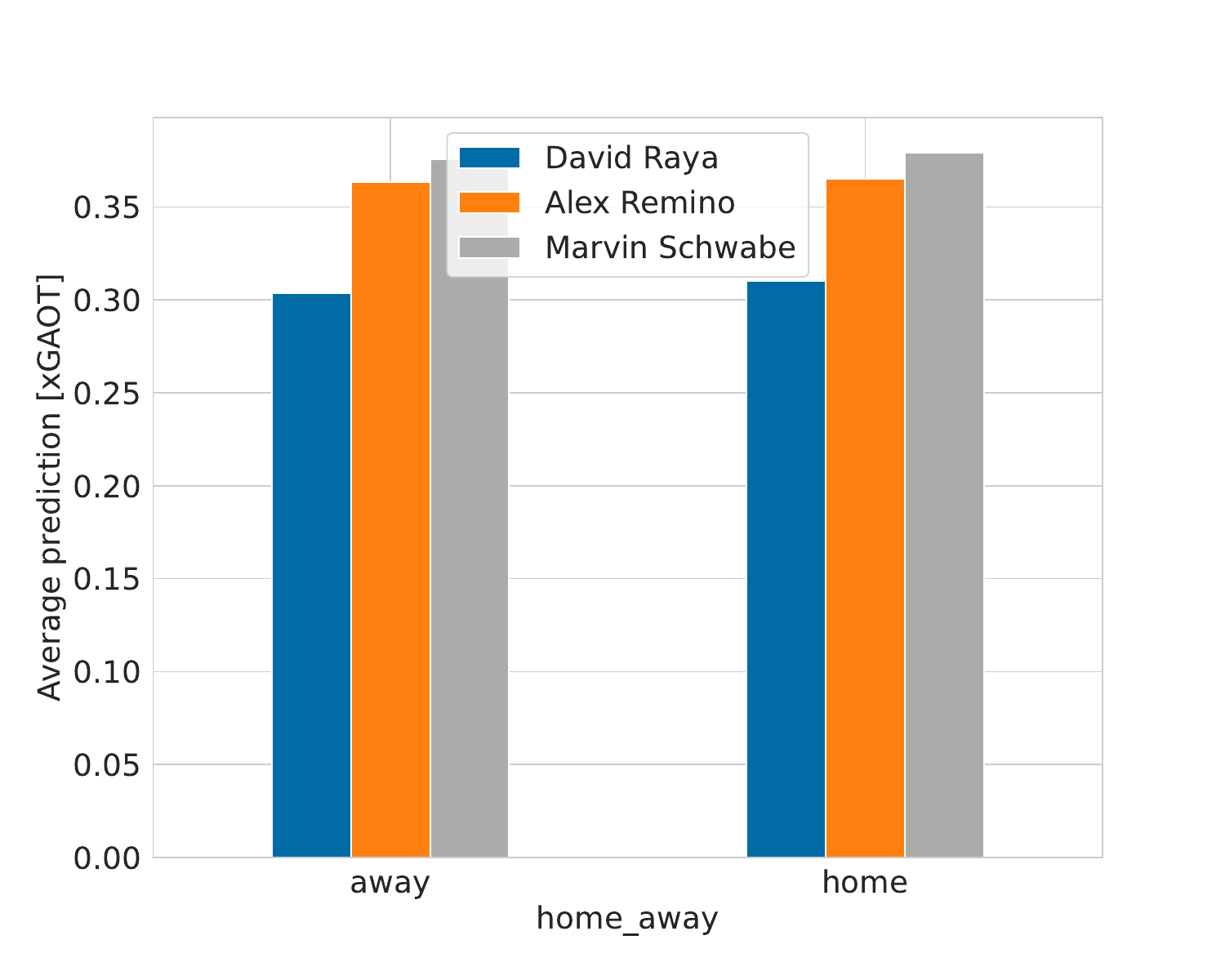}
    \caption{Aggregated goalkeeper blind spot profiles for the variable \texttt{homeAwayTeam}.}
    \label{fig:gk-ap-h_a}
\end{figure}

The aggregated profiles of the goalkeepers for the variables \texttt{situation}, \texttt{shot type}, and \texttt{home and away} are given in Fig~\ref{fig:gk-ap-situation}, \ref{fig:gk-ap-shot}, \ref{fig:gk-ap-h_a}, respectively. In these figures, the y-axis represents the average predictions of the xG model for \textbf{xGAOT}.

In Fig~\ref{fig:gk-ap-situation}, we can compare the goalkeepers' blind spots in relation to the situation in the match. It is seen that the harder situations are \textit{From Corner}, \textit{Set Piece}, \textit{Direct Freekick}, respectively. \textbf{Marvin Schwabe} shows better performance in all situations except \textit{Set Piece}. It can be evaluated that it is a blindspot in his performance when he is compared with others. A similar finding is captured from Fig~\ref{fig:gk-ap-shot}. Again \textbf{Marvin Schwabe} shows better performance against all shot types except \textit{Head}. It is known that playing at home is slightly motivating for the players (see Fig~\ref{fig:gk-ap-h_a}). There are no significant differences between the performance of goalkeepers at home and away. As a result, \textbf{Marvin Schwabe} shows better performance than others in terms of aggregated profiles for the variables considered.

\subsection{Performance change analysis with the aSHAP} \label{sec:performance_change}

This section shows how aSHAP values can be used to compare the team's performance across the selected seasons. We focused on the performance of the teams \textbf{SSC Napoli} and \textbf{Lille OSC} because their performance has changed drastically positively and negatively, respectively in recent years. \textbf{SSC Napoli} became the league champion in the season of 2022/23 while they finished the league in 3rd place in the previous season, and \textbf{Lille OSC} became the league champion in the season 2020/21 but finished the league in 10th place the next season.

\begin{figure}
\centering
\begin{subfigure}{.70\textwidth}
  \centering
    \includegraphics[width=\textwidth]{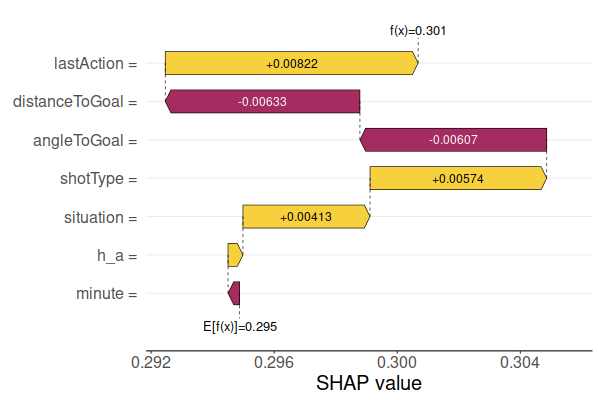}
    \caption{The season of 2021/22}
    \label{fig:napoli-2021}
\end{subfigure} \hfill
\begin{subfigure}{.70\textwidth}
  \centering
    \includegraphics[width=\textwidth]{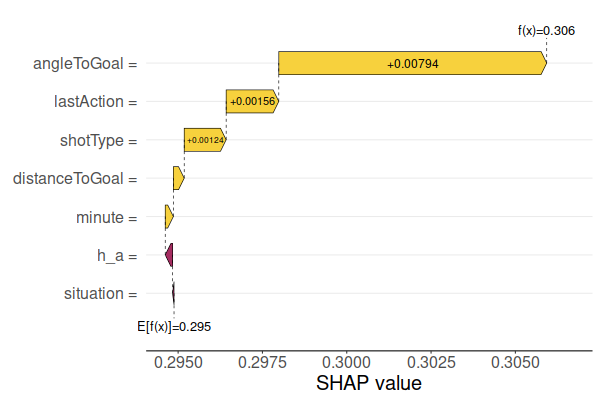}
    \caption{The season of 2022/23}
    \label{fig:napoli-2022}
\end{subfigure}
\caption{The aSHAP profiles for \textbf{SSC Napoli} in the seasons of 2021/22 and 2022/23.}
\label{fig:napoli}
\end{figure}

Here we decomposed the contribution of the variables in terms of the aSHAP values during the selected seasons, and then we compared the values to capture valuable insights into the reason for the drastic change in the team performance. The aSHAP profiles of the \textbf{SSC Napoli}, which is the first example in this section, during the seasons of 2021/22 and 2022/23 are given in Fig.~\ref{fig:napoli}. It is seen that the contribution of the variables \texttt{distanceToGoal} and \texttt{angleToGoal} is negative to the xG model in the season of 2021/22 (\textit{the worse season}) in Fig~\ref{fig:napoli-2021}, while it is positive in the season of 2022/23 (\textit{the better season}) as seen in Fig~\ref{fig:napoli-2022}. These are the most important variables in the xG models, which means that changes in these variables have the greatest impact on the average model prediction. From the aSHAP profiles, It may be concluded that the changes made to the variables with negative contributions increased the team's performance.

\begin{figure}
\centering
\begin{subfigure}{.7\textwidth}
  \centering
    \includegraphics[width=\textwidth]{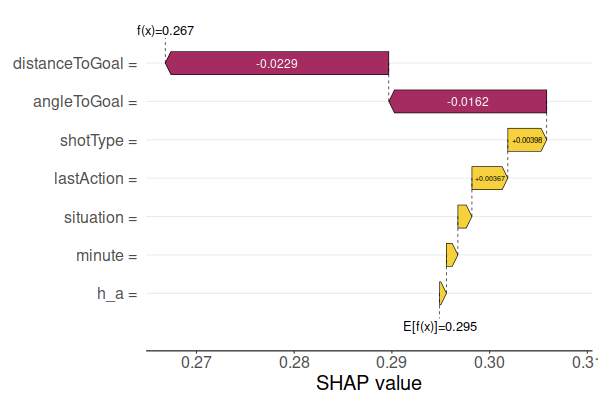}
    \caption{The season of 2020/21}
    \label{fig:lille-2020}
\end{subfigure} \hfill
\begin{subfigure}{.7\textwidth}
  \centering
    \includegraphics[width=\textwidth]{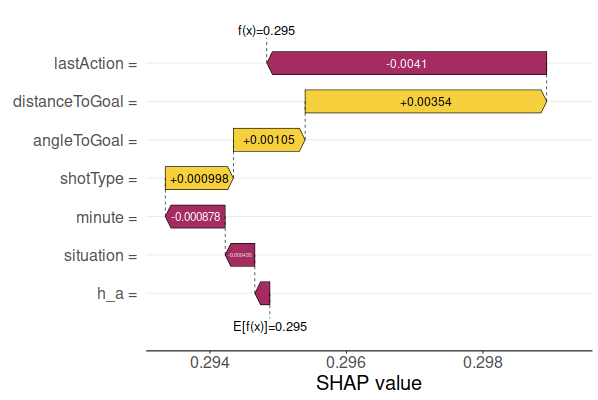}
    \caption{The season of 2021/22}
    \label{fig:lille=2021}
\end{subfigure}
\caption{The aSHAP profiles for \textbf{Lille OSC} in the season of 2020/21 and 2021/22.}
\label{fig:lille}
\end{figure}

Another example is \textbf{Lille OSC} from Ligue 1 and its aSHAP profiles during the seasons of 2020/21 and 2022/22 are given in Fig.~\ref{fig:lille}. One of the interesting insights about the performance of \textbf{Lille OSC} is they have a lower xG per shot of $0.267$ during the season of 2020/21 than $0.295$ in their worse season. Since their success or failure at the end of the season is not only related to their performance, the emergence of this situation can also be interpreted as normal. However, it can be concluded that their performance may be worsened because of the decreasing contribution of the \texttt{lastAction}. It contributed negatively to the average xG as $-0.0041$ and this is the highest effect on the xG during the worse season of \textbf{Lille OSC}.

The performance of a team or a player can be explained in terms of aSHAP values to decompose the contribution of variables in the xG model. Moreover, the changes in the performance of them can be analyzed by comparing the different periods. Thus, the reason(s) responsible for the change of interest can be obtained by following this approach.

\section{Conclusions} \label{sec:conclusions}

In this paper, we propose the use of glocal explanations of xG models in performance analysis. We introduced the aggregated SHAP values and highlighted the aggregated profiles as glocal-level XAI tools. These methodologies provided valuable insights into player and team performance instead of a single shot (e.g., a single observation). We show that the aggregated SHAP is practically effective in understanding the contribution of individual variables to the xG model's predictions. By decomposing the model's predictions into individual components, we gained a clear understanding of the impact of each variable on the model's overall prediction. This information allowed us to assess the performance of a player or team over a period of time. 

Furthermore, the aggregated profiles provided a useful tool for comparing players' performance and identifying strengths and weaknesses in terms of scoring potential. We identified young players with high-scoring potentials, such as Evan Ferguson and Mathys Tel, by analyzing their aggregated profiles. They are also applied to analyze the blind spot in goalkeepers' performance. By examining the profiles for variables such as situation, shot type, and home and away matches, we gained insight into goalkeepers' performance under different conditions. Marvin Schwabe showcased better performance in various situations and shot types, except for Set Piece situations, which revealed a blind spot in his performance. These findings demonstrated the potential of aggregated profiles in evaluating and comparing goalkeeper performance.

Our study highlights the usefulness of glocal explanations in soccer performance analysis. By incorporating both local and global perspectives, we were able to gain comprehensive insights into player and team performance. The glocal explanations provided valuable information for evaluating scoring potential, identifying blind spots, and making data-driven decisions in player selection and tactical strategies. However, it is important to note that our study focused on specific variables and datasets. Further research is needed to explore additional variables and consider larger datasets for a more comprehensive analysis. Additionally, the application of glocal explanations can be extended to other areas of soccer analytics, such as defensive performance analysis and team-level evaluations. We emphasize that the limitation of aSHAP is its heavy computational steps. Further research can be aimed at handling this issue. In conclusion, the application of glocal explanations, specifically aggregated SHAP values and aggregated profiles, provides valuable insights into xG models in soccer performance analysis. These methodologies offer a deeper understanding of player contributions, scoring potential, and blind spots in goalkeeper performances. The findings from this study contribute to the growing field of soccer analytics and pave the way for further advances in performance analysis and decision-making processes in soccer.

\section*{Declarations}

\noindent \textbf{Author contributions } All authors contributed to the study conception and design. Data collection, analysis, and visualizations were performed by Adrian Stańdo. The original draft of the manuscript was written by Mustafa Cavus and all authors commented on previous versions of the manuscript. All authors read and approved the final manuscript. Przemysław Biecek contributed to providing feedback and guidance.\\

\noindent \textbf{Funding } The work on this paper is financially supported by the NCN Sonata Bis-9 grant 2019/34/E/ST6/00052.  This work was carried out with the support of the Laboratory of Bioinformatics and Computational Genomics and the High-Performance Computing Center of the Faculty of Mathematics and Information Science, Warsaw University of Technology under computational grant number A-22-09.  \\

\noindent \textbf{Acknowledgements } We would like to thank Mikołaj Spytek and Katarzyna Woźnica for their useful comments.\\

\noindent \textbf{Conflict of interest } Not applicable.\\

\noindent \textbf{Ethics approval } Not applicable.\\

\noindent \textbf{Consent to participate } Not applicable.\\

\noindent \textbf{Consent for publication } All authors consent to submission and publication.\\

\noindent \textbf{Availability of data and materials } This research does not contribute any new data. All data used is existing and publicly available.\\

\noindent \textbf{Code availability } The codes needed to reproduce the results can be found in the following GitHub repository: \url{https://github.com/adrianstando/glocal-explanations-of-xG-models}.\\

\bibliography{sn-bibliography}

\end{document}